\documentclass[aps,pre,floats,twocolumn,superscriptaddress]{revtex4}

\pdfoutput=1

\usepackage{latexsym,amsmath}
\usepackage{amsbsy}
\usepackage[usenames]{color}
\usepackage{amssymb}
\usepackage{graphicx}
\usepackage{url}
\usepackage{alltt}

\begin{document}

\title{Beyond Zipf's law:\\
Modeling the structure of human language}

\author{M. \'{A}ngeles Serrano}
\affiliation{IFISC (CSIC-UIB) Instituto de F\'isica
Interdisciplinar y Sistemas Complejos,\\
Campus Universitat Illes Balears, E-07122 Palma de Mallorca, Spain}

\author{Alessandro Flammini}
\affiliation{School of Informatics, Indiana University, Bloomington, USA}

\author{Filippo Menczer}
\affiliation{School of Informatics, Indiana University, Bloomington, USA} \affiliation{Complex Networks Lagrange Lab, ISI Foundation, Torino, Italy}

%\date{}

\begin{abstract}
Human language, the most powerful communication system in history, is closely associated with cognition. Written text is one of the fundamental manifestations of language, and the study of its universal regularities can give clues about how our brains process information and how we, as a society, organize and share it. Still, only classical patterns such as Zipf's law have been explored in depth. In contrast, other basic properties like the existence of bursts of rare words in specific documents, the topical organization of collections, or the sublinear growth of vocabulary size with the length of a document, have only been studied one by one and mainly applying heuristic methodologies rather than basic principles and general mechanisms. As a consequence, there is a lack of understanding of linguistic processes as complex emergent phenomena. Beyond Zipf's law for word frequencies, here we focus on Heaps' law, burstiness, and the topicality of document collections, which encode correlations within and across documents absent in random null models. We introduce and validate a generative model that explains the simultaneous emergence of all these patterns from simple rules. As a result, we find a connection between the bursty nature of rare words and the topical organization of texts and identify dynamic word ranking and memory across documents as key mechanisms explaining the non trivial organization of written text. Our research can have broad implications and practical applications in computer science, cognitive science, and linguistics.
\end{abstract}

\maketitle

\section*{Introduction}

Even in the era of the information technology revolution,
language remains the most powerful and sophisticated communication system in the history of civilization~\cite{Hauser:2002}. Its understanding requires an interdisciplinary approach and has broad conceptual and practical implications. It involves a number of disciplines; from computer science, where natural language processing~\cite{Joshi:1991,ManningSchutze99,Nowak02Computatio} seeks to model language computationally, to cognitive science, that tries to understand our intelligence with linguistics as one of its key contributing disciplines~\cite{Chomsky:2006}.

After speech, written text is probably the most fundamental manifestation of human language. Nowadays, electronic and information technology media offer the opportunity of recording and accessing easily huge amounts of documents that can be analyzed in quest for some of the signatures of human communication. As a first step, statistical patterns in written text can be detected as a trace of the mental processes we use in communication. It has been realized that various universal regularities characterize text from different domains and languages. The best-known is Zipf's law on the distribution of word frequencies~\cite{Zipf49Human-Beha,Baayen01Word-Frequ,Saichev:2008}, according to which the frequency of terms in a collection decreases inversely to the rank of the terms. Zipf's law has been found to apply to collections of written documents in virtually all languages. Other notable universal regularities of text are Heaps' law~\cite{Heaps78Informatio,cattuto_pnas07_tagging}, according to which vocabulary size grows slowly with document size, i.e. as a sublinear function of the number of words; and the bursty nature of words~\cite{Church95Poisson-mi,Katz96Distributi,Kleinberg02bursty}, making a word more likely to reappear in a document if it has already appeared, compared to its overall frequency across the collection.

Understanding the structure of written text is key to a broad range of critical applications such as Web search~\cite{Chakrabarti03,LiuWebMiningBook} (and the booming business of online advertising), literature mining~\cite{05Text-Minin,Feldman06The-Text-M},
topic detection~\cite{Allan98tdt,Yang98tdt}, and security~\cite{Chen06ISI,Newman06Analyzing, Pennebaker07}. Thus, it is not surprising that researchers in linguistics, information and cognitive science, machine learning, and complex systems are coming together to model how universal text properties emerge. Different models have been proposed that are able to predict each of the universal properties outlined above. However, no single model of text generation explains all of them together. Furthermore, no model has been used to interpret or predict the empirical distributions of text similarity between documents in a collection~\cite{Menczer03attach,Menczer02navigation}.
\begin{figure*}[t]
\centerline{\includegraphics[width=0.9\textwidth]{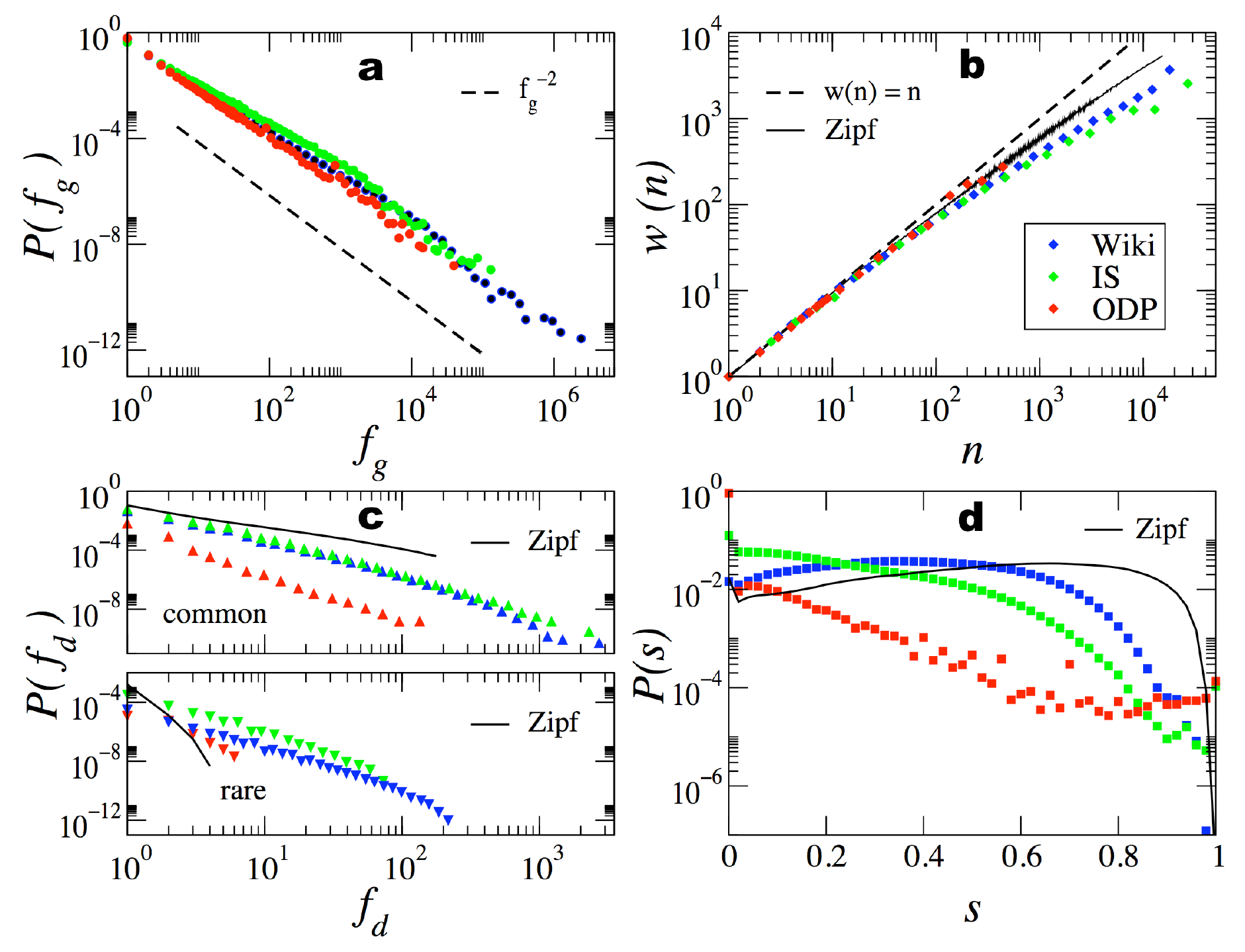}}
\caption{Regularities in textual data as observed in our three empirical datasets.
(a) Zipf's Law: word counts are globally distributed according to a power law $P(f_g) \sim f_g^{-2}$. (b) Heaps' Law: as the number of words $n$ in a document grows, the average vocabulary size (i.e. the number of distinct words) $w(n)$ grows sublinearly with $n$. (c) Burstiness: fraction of documents $P(f_d)$ containing $f_d$ occurrences of common or rare terms. For each dataset, we label as ``common'' those terms that account for 71\% of total word occurrences in the collection, while rare terms account for 8\%.
(d) Similarity: distribution of cosine similarity $s$ across all pairs of documents, each represented as a term frequency vector. Also shown are $w(n)$, the distributions of $f_d$, and the distribution of $s$ according to the Zipf null model (see text) corresponding to the IS dataset.}
\label{empirical}
\end{figure*}

In this paper we present a model that generates collections of documents consistently with all of the above statistical features of textual corpora, and validate it against large and diverse Web datasets. We go beyond the global level of Zipf's law, which we take for granted, and focus on general correlation signatures within and across documents. These correlation patterns, manifesting themselves as burstiness and similarity, are destroyed when the words in a collection are reshuffled, even while the global word frequencies are preserved. Therefore the correlations are not simply explained by Zipf's law, and are directly related to the global organization and topicality of the corpora. The aim of our model is not to reproduce the microscopic patterns of occurrence of individual words, but rather to provide a stylized generative mechanism to interpret their emergence in statistical terms. Consequently, our main assumption is a global distribution of word probabilities; we do not need to fit a large number of parameters to the data, in contrast to parametric models proposed to describe the bursty nature or topicality of text~\cite{Blei03,Steyverso4topics,Elkan06Clustering}. In our model, each document is derived by a local ranking of dynamically reordered words, and different documents are related by sharing subsets of these rankings that represent emerging topics. Our analysis shows that the statistical structure of text collections, including their level of topicality, can be derived from such a simple ranking mechanism. Ranking is an alternative to preferential attachment for explaining scale invariance~\cite{Goh:2001} and has been used to explain the emergent topology of complex information, technological, and social networks~\cite{Fortunato05ranking}. The present results suggest that it may also shed light on cognitive processes such as text generation and the collective mechanisms we use to organize and store information.

\section*{Empirical observations}

We have selected three very diverse public datasets, from topically focused to broad coverage, to illustrate the statistical regularities of text and validate our model. The first corpus is the Industry Sector database (IS), a collection of corporate Web pages organized into categories or sectors. The second dataset is a sample of the Open Directory (ODP), a collection of Web pages classified into a large hierarchical taxonomy by volunteer editors. The third corpus is a random sample of topic pages from the English Wikipedia (Wiki), a popular collaborative encyclopedia which also is comprised of millions of online entries. (See Appendix A for details.)

We measured the statistical regularities mentioned above in our datasets and the empirical results are shown in Fig.~1. The distributions of document length for all three collections is very well approximated by a log-normal, with different first and second moment parameters (see Table~1 and Fig.~1 in Appendix A). According to Zipf's law~\cite{Zipf49Human-Beha,Baayen01Word-Frequ,VP-Maslov06On-Zipfs-l,Saichev:2008}, the global frequency $f_g$ of terms in a collection decreases roughly inversely to their rank $r$: $f_g \sim 1/r$ or, in other words, the distribution of the frequency $f_g$ is well approximated by a power law $P(f_g) \sim f_g^{-\alpha}$ with exponent $\alpha \approx 2$. This regularity has been found to apply to collections of written documents in virtually all languages, including the datasets used here (Fig.~1a). Heaps' law~\cite{Heaps78Informatio,cattuto_pnas07_tagging} describes the sublinear growth of vocabulary size (number of unique words) $w$ as a function of the size of a document (number of words) $n$ (Fig.~1b). This regularity has also been observed in different languages, and the behavior has been interpreted as a power law $w(n) \sim n^{\beta}$ with $\beta < 1$, although the exponent $\beta$ between 0.4 and 0.6 is language-dependent~\cite{baeza}.

Burstiness is the tendency of some words to occur clustered together in individual documents, so that a term is more likely to reappear in a document where it has appeared before~\cite{Church95Poisson-mi,Katz96Distributi,Kleinberg02bursty}. This property is more evident among rare words, which are more likely to be topical. Following Elkan~\cite{Elkan06Clustering}, the bursty nature of words can be illustrated by dividing words into classes according to their global frequency (e.g., common vs. rare). For words in each class, we plot in Fig.~1c the fraction $P(f_d)$ of documents in which these words occur with frequency $f_d$. We compare the distribution $P(f_d)$ of common and rare terms with those predicted by the null independence hypothesis, that generates documents whose length is drawn from the same lognormal distribution as the empirical data (see Table~1 and Fig.~1 in Appendix A) by drawing words independently at random from the global Zipf frequency distribution (Fig.~1a). As compared to the reference of such a \emph{Zipf model}, rare terms are much more likely to cluster in specific documents and not to appear evenly distributed in the collection, so that ordering principles beyond those responsible for Zipf's law have to be at play.

Another signature of text collections, which is more telling about topicality, is the distribution of lexical similarity across pairs of documents. In information retrieval and text mining, documents are typically represented as term vectors~\cite{Salton83,LiuWebMiningBook}. Each element of a vector represents the weight of the corresponding term in the document. There are various vector representations according to different weighting schemes. Here, we focus on the simplest scheme, in which a weight is simply the frequency of the term in the document. The similarity between two documents is given by the cosine between the two vectors:
$s(p,q) = \sum_t w_{tp} w_{tq} / \sqrt{\sum_t w_{tp}^2 \cdot \sum_t w_{tq}^2}$, where $w_{tp}$ is the weight of term $t$ in document $p$. It has been observed that for documents sampled from the ODP, the distribution of cosine similarity based on term frequency vectors is concentrated around zero and decays in a roughly exponential fashion for $s>0$~\cite{Menczer02navigation,Menczer03attach}. Figure~1d shows that different collections yield different similarity profiles, however they all tend to be more skewed toward small similarity values than predicted by the Zipf model.

Modeling how these properties emerge from simple rules is central to an understanding of human language and related cognitive processes. Our understanding, however, is far from definitive. First, because the empirical observations are open to different interpretations. As an example, much has been written about the debate between Simon and Mandelbrot around different interpretations of Zipf's law (see \url{www.nslij-genetics.org/wli/zipf} for a historical review of the debate). Second, and perhaps more importantly, no single model of text generation explains all of the above observations simultaneously. Third, models at hand are usually based on heuristic methods rather than on basic principles and general mechanisms that could explain linguistic processes as emergent phenomena.

In the remainder of this paper, we focus on burstiness and similarity distributions. Regarding similarity, little attention has been given to its empirical distribution and, to the best of our knowledge, no model has been put forth to explain its profile. Regarding text burstiness, on the other hand, several models have been proposed including the two-Poisson \linebreak model~\cite{Church95Poisson-mi}, the Poisson zero-inflated mixture model~\cite{Jansche03}, Katz' k-mixture model~\cite{Katz96Distributi}, and a gap-based variation of Bayes model~\cite{Sarkar05}. Another line of generative models extends the simple multinomial family with increasingly complex views of topics. Examples include probabilistic latent semantic indexing~\cite{Hofmann99}, latent Dirichlet allocation (LDA)~\cite{Blei03}, and Pachinko allocation~\cite{McCallum06pachinko}. These models assume a set of topics, each typically described by a multinomial distribution over words. Each document is then generated from some mixture of these topics. In LDA, for example, the parameters of the mixture are drawn from a Dirichlet distribution, independently for each document. Each word in a document is generated by drawing a topic from the mixture and then the term from its corresponding word distribution. A variety of techniques have been developed to estimate from data the parameters that characterize the many distributions involved in the generative process~\cite{Steyverso4topics,griffiths05,Newman06Analyzing}. Although the above models were mainly developed for subject classification, they have also been used to investigate burstiness since bursty words can characterize the topic of a document~\cite{Elkan05,Elkan06Clustering}.

The very large numbers of free parameters associated with individual terms, topics, and/or their mixtures grant the above models great descriptive power. However, their cognitive plausibility is problematic. Our aim here is instead to produce a simpler, more plausible mechanism compatible with the high-level statistical regularities associated with \emph{both} burstiness and similarity distributions, without regard for explicit topic modeling.

\section*{Model and results}

Two basic mechanisms, reordering and memory, can explain burstiness and similarity consistently with Zipf's law. We show this by proposing a generative model that incorporates these processes to produce collections of documents characterized by the observed statistical regularities. Each document is derived by a local ranking of words that reorganizes according to the changing word frequencies as the document grows, and different documents are related by sharing subsets of these rankings that represent emerging topics. With just the main assumptions of the global distribution of word probabilities and document sizes and a single tunable parameter measuring the topicality of the collection, we are able to generate synthetic corpora that re-create faithfully the features of our Web datasets. Next, we describe two variations of the model, one without memory and the second with a memory mechanism that captures topicality.
\begin{figure*}[t]
\centerline{\includegraphics[width=1\textwidth]{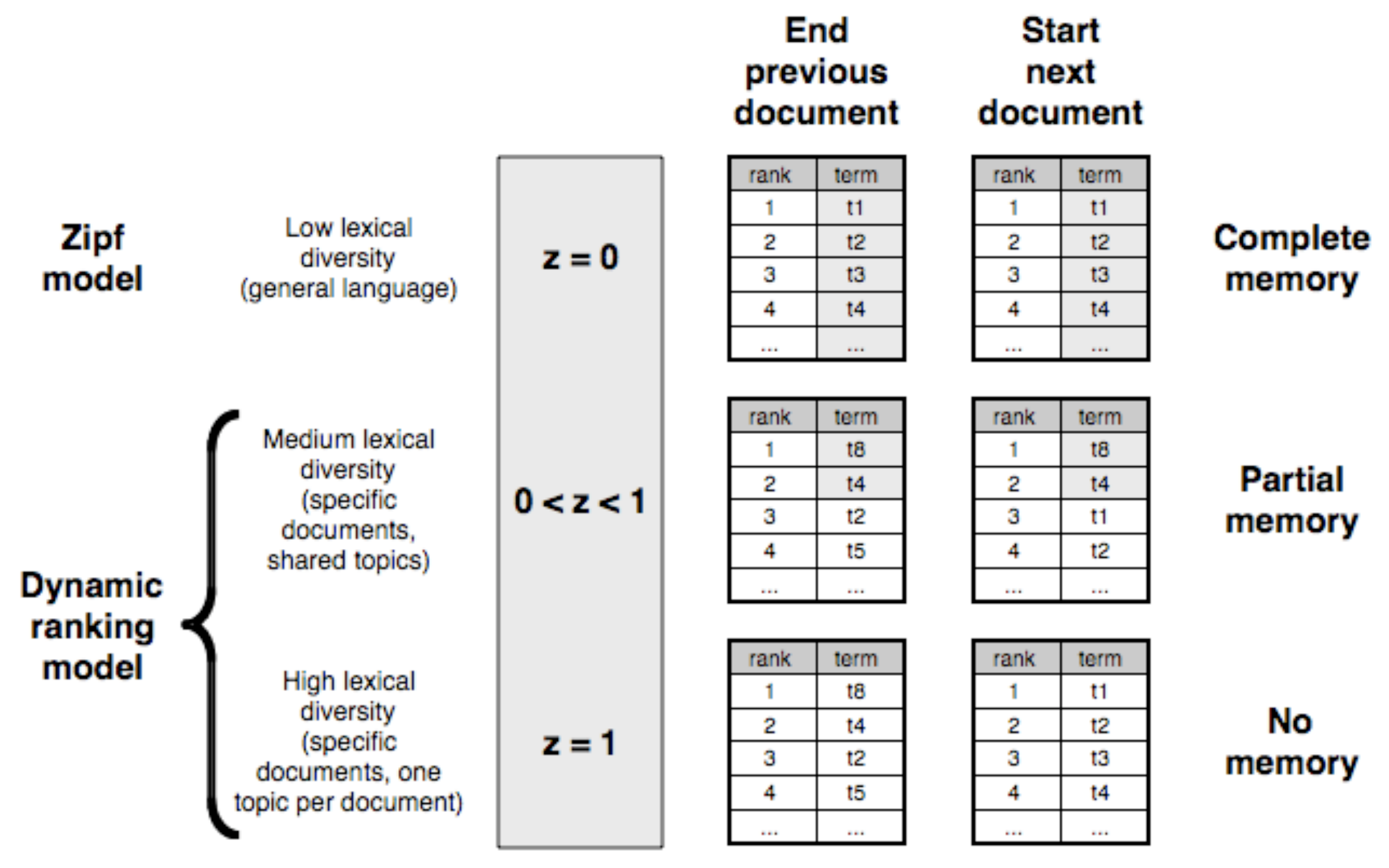}}
\caption{Illustration of the dynamic ranking model. The parameter $z$ regulates the lexical diversity, or topicality of the collection. The extreme case $z=0$ is equivalent to the null Zipf model, where all documents are generated using the global word rank distribution. The opposite case $z=1$ is the first version of the dynamic ranking model, with no memory, in which each new document starts from the global word ranking $r_0$. Intermediate values of $z$ represent the more general version of the dynamic ranking model, where correlations across documents are created by a partial memory of word ranks. A more detailed algorithmic description of the model can be found in Appendix C.}
\label{fig:algo}
\end{figure*}

\subsection*{Dynamic ranking by frequency}

In our model, $D$ documents are generated drawing word instances repeatedly with replacement from a vocabulary of $V$ words. The document lengths in number of words are drawn from a lognormal distribution. The parameters $D$, $V$ and the lognormal mean and variance are derived empirically from each dataset (see Table~1 in Appendix A). We further assume that at any step of the generation process, word probabilities follow a Zipf distribution $P[r(t)] \propto r(t)^{-1}$ where $r(t)$ is the rank of term $t$. However, rather than keeping a fixed ranking, we imagine that words are sorted dynamically during the generation of each document according to the number of times they have already occurred. Words and ranks are thus decoupled: at different times, a word can have different ranks and a position in the ranking can be occupied by different words. The idea is that as the topicality of a document emerges through its content, topical words will be more likely to reoccur within the same document. This idea is incorporated into the model as a frequency bias favoring words that occur early in the document.

In the first version of the model, each document is produced independently of each other. Before each new document is generated, words are sorted according to an initial global ranking, which remains fixed for all documents. This ranking $r_0$ is also used to break ties during the generation of documents, mong words with the same occurrence counts. The algorithm corresponding to this dynamic ranking model is illustrated in Fig.~2 and detailed in Appendix C.

When a sufficiently large number of documents is generated,
the measured frequency of a word $t$ over the entire corpus approaches the Zipf distribution $P(t) \sim [r_0(t)]^{-1}$, ensuring the self consistency of the model. We numerically simulated the dynamic ranking model for each dataset.
A direct comparison with the empirical burstiness curves shown in Fig.~1c can be found in Fig.~3a. The excellent agreement
suggests that the dynamic ranking process is sufficient for producing the right amount of correlations inside documents needed to realistically account for the burstiness effect.
\begin{figure*}[t]
\centerline{\includegraphics[width=1\textwidth]{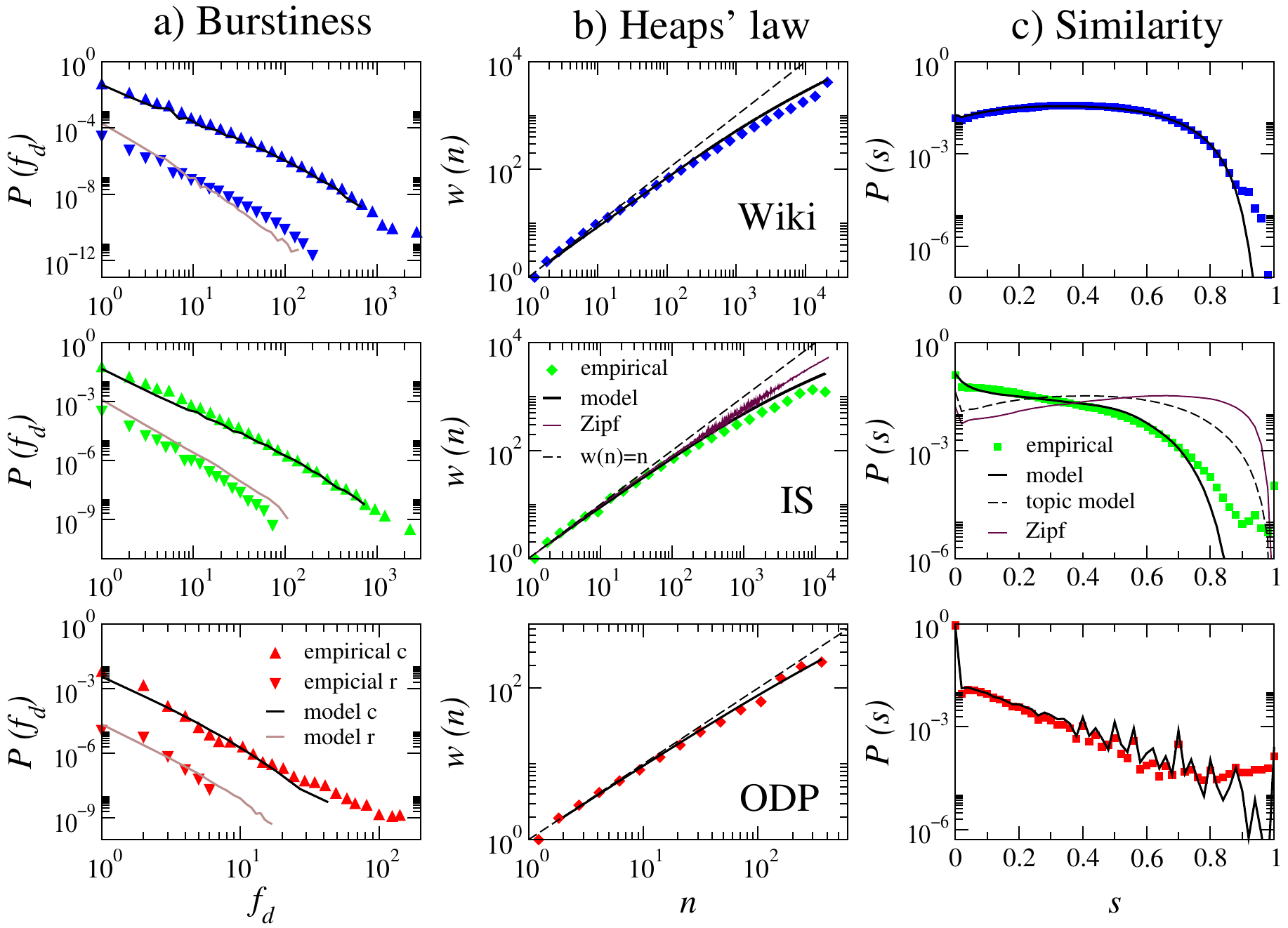}}
\caption{Model vs.~empirical observations. (a) Comparison of burstiness curves produced by the dynamic ranking model with those from the empirical datasets. Common and rare words are defined in Fig.~1c. (b) Comparison of Heaps' law curves produced by the dynamic ranking model with those from the empirical datasets. Simulations of the model provide the same predictions as numerical integration of the analytically derived equation using the empirical rank distributions (see Appendix B). For the IS dataset we also plot the result of the Zipf null model, which produces a sublinear $w(n)$, although less pronounced than our model. The ODP collection has short documents on average (cf.~Table~1 in Appendix A), so Heaps' law is barely observable. (c) Comparison between similarity distributions produced by the dynamic ranking model with memory, and those from the empirical datasets also shown in Fig.~1d. The parameter $z$ controlling the topical memory is fitted to the data. The peak at $s=0$ suggests that the most common case is always that of documents sharing very few or no common terms. The discordance for high similarity values is due to corpus artifacts such as mirrored pages, templates, and very short (one word) documents. The fluctuations in the curves for the ODP dataset are due to binning artifacts for short pages. Also shown is the prediction of the topic model for the IS dataset (see text).}
\label{fig:model}
\end{figure*}

Heaps' law can be derived analytically from our model (see Appendix B). Assuming a Zipf's law with a tail of the form $P(r) \sim r^{-\gamma}$ where $\gamma > 1$, the solution is $w(n) \sim n^{1/\gamma}$ and we recover Heaps' sublinear growth with $\beta \approx 1/\gamma$ for large $n$. According to the Yule-Simon model~\cite{yule-simon}, which interprets Zipf's law through a preferential attachment process, the rank distribution should have a tail with exponent $\gamma > 1$. This is confirmed empirically in many English collections; for example our ODP and Wikipedia datasets yield Zipfian tails with $\gamma$ between 3/2 and 2. Our model predicts that in these cases Heaps' growth should be well approximated by a power law with exponent $\beta$ between 1/2 and 2/3, closely matching those reported for the English language~\cite{baeza}. Simulations using the empirically derived $P(r)$ for each dataset display growth trends for large $n$ that are in good agreement with the empirical behavior (Fig.~3b).

\subsection*{Topicality and similarity}

The agreement between empirical data and simulations of the model with respect to the similarity distributions gets worse for those datasets that are more topically focused. A new mechanism is needed to account for topical correlations between documents.

The model in the previous section generates collections of independent text documents, with specific but uncorrelated topics captured by the bursty terms. For each new document, the rank of each word $t$ is initialized to its original value $r_0(t)$ so that each document has no bias toward any particular topic. The synthetic corpora which result display broad coverage. However, real corpora may cover more or less specific topics. The stronger the semantic relationship between documents, the higher the likelihood they share common words. Such collection topicality needs to be taken into account to accurately reproduce the distribution of text similarity between documents.

To incorporate topical correlations into our model, we introduce a memory effect connecting word frequencies across different documents. Generative models with memory have already been proposed to explain Heaps' law~\cite{cattuto_pnas07_tagging}. In our algorithm (see Fig.~2 and Appendix C) we replace the initialization step so that a portion of the initial ranking of the terms in each document is inherited from the previously generated document. In particular, the counts of the $r^*$ top-ranked words are preserved while all the others are reset to zero.
The rank $r^{*}$ is drawn from an exponential distribution $P(r^*) = z(1-z)^{r^* - 1}$ where $z$ is a probability parameter that models the lexical diversity of the collection and $r^*$ has expected value $1/z$, which can be interpreted as the collection's shared topicality.

This variation of the model does not interfere with the reranking mechanism described in the previous section, so that the burstiness effect is preserved. The idea is to interpolate between two extreme cases. The case $z=0$, in which counts are never reset, converges to the null Zipf model. All documents share the same general terms, modeling a collection of unspecific documents. Here we expect a high similarity in spite of the independence among documents, because the words in all documents are drawn from the identical Zipf distribution. The other extreme case, $z=1$, reduces to the original model, where all the counts are always initialized to zero before starting a document. In this case, the bursty words are numerous but not the same across different documents, modeling a situation in which each document is very specific but there is no shared topic across documents. Intermediate cases $0<z<1$ allow us to model correlations across documents not only due to the common general terms, but also to topical (bursty) terms.

We simulated the dynamic ranking model with memory under the same conditions corresponding to our datasets, but additionally fitting the parameter $z$ to match the empirical similarity distributions. The comparisons are shown in Fig.~3c. The similarity distribution for the ODP is best reproduced for $z=1$, in accordance to the fact that this collection is overwhelmingly composed of very specific documents spanning all topics. In such a situation, the original model accurately reproduces the high diversity among document topics and there is no memory need. In contrast, Wikipedia topic pages use a homogenous vocabulary due to their strict encyclopedic style and the social consensus mechanism driving the generation of content. This is reflected in the value $z=0.005$, corresponding to an average of $1/z = 200$ common words whose frequencies are correlated across successive pairs of documents. The industry sector dataset provides us with an intermediate case in which pages deal with more focused, but semantically related topics. The best fit of the similarity distribution is obtained for $z=0.1$.

With the fitted values for the shared topicality parameter $z$, the agreement between model and empirical similarity data in Fig.~3c is excellent over a broad range of similarity values.
To better illustrate the significance of this result, let us compare it with the prediction of a simple topic model. For this purpose one must have a priori knowledge of a set of topics to be used for generating the documents. The IS dataset lends itself to this analysis because the pages are classified into twelve disjoint industry sectors, which can naturally be interpreted as unmixed topics. For each topic $c$, we measured the frequency of each term $t$ and used it as a probability $p(t|c)$ in a multinomial distribution. We generated the documents for each topic using the actual empirical values for the number of documents in the topic and the number of words in each document. As shown in Fig.~3c, the resulting similarity distribution is better than that of the Zipf model (where we assume a single global distribution), however the prediction is not nearly as good as that of our model.

Our model only requires a single free parameter $z$ plus the global (Zipfian) distribution of word probabilities, which determines the initial ranking. Conversely, for the topic model we must have ---or fit--- the frequency distribution $p(t|c)$ over all terms for each topic, which implies an extraordinary increase in the number of free parameters since, apart from potential differences in the functional forms, each distribution would rank the terms in a different order.

Aside from complexity issues, the ability to recover similarities suggests that the dynamic ranking model, though not as well informed as the topic model on the distributions of the specific topics, better captures word correlations. Topics emerge as a consequence of the correlations between bursty terms across documents as determined by $z$, but it is not necessary to predefine explicitly the number of topics or their distributions as other models require.

\section*{Conclusion}

Our results show that key regularities of written text beyond Zipf's law, namely burstiness, topicality and their interrelation, can be accounted for on the basis of two simple mechanisms, namely frequency ranking with dynamic reordering and memory accross documents, and can be modeled with an essentially parameter-free algorithm. The rank based approach is in line with other recent models in which ranking has been used to explain the emergent topology of complex information, technological, and social networks~\cite{Fortunato05ranking}. It is not the first time that a generative model for text has walked parallel paths with models of network growth. A remarkable example is the Yule-Simon model of text generation~\cite{yule-simon}, which was later rediscovered in the context of citation analysis~\cite{deSollaPrice76}, and has recently found broad popularity in the complex networks literature~\cite{Albert02}.

Our approach applies to datasets where the temporal sequence of documents is not important, but burstiness has also been studied in contexts where time is a critical component~\cite{Kleinberg02bursty,Barabasi05bursty}, and even in human languages evolution~\cite{Atkinson:2008}. Further investigations in relation to topicality could attempt to explicitly demonstrate the role of the topicality correlation parameter by looking at the hierarchical structure of content classifications. Subsets of increasingly specific topics of the whole collection could be extracted to study how the parameter $z$ changes and how it is related to external categorizations. The proposed model can also be used to study the coevolution of content and citation structure in the scientific literature, social media such as the Wikipedia, and the Web at large~\cite{Kleinberg04Analysing,Menczer03attach,Alvarez-Lacalle06Hierarchic,cattuto_pnas07_tagging}.

From a broader perspective, it seems natural that models of text generation should be based on similar cognitive mechanisms as models of human text processing since text production is a translation of semantic concepts in the brain into external lexical representations. Indeed, our model's connection between frequency ranking and burstiness of words provides a way to relate two key mechanisms adopted in modeling how humans process the lexicon: rank frequency~\cite{Murray04} and context diversity~\cite{Adelman06}. The latter, measured by the number of documents that contain a word, is related to burstiness since given a term's overall collection frequency, higher burstiness implies lower context diversity. While tracking frequencies is a significant cognitive burden, our model suggests that simply recognizing that a term occurs more often than another in the first few lines of a document would suffice for detecting bursty words from their ranking and consequently the topic of the text.

In summary, a picture of how language structure and topicality emerge in written text as complex phenomena can shed light into the collective cognitive processes we use to organize and store information, and find broad practical applications, for instance, in topic detection, literature analysis, and Web mining.

\begin{acknowledgments}
We thank Charles Elkan, Rich Shiffrin, Michael Jones, Alessandro Vespignani, Dragomir Radev, Vittorio Loreto, Ciro Cattuto, and Mari{\'a}n Bogu{\~n}{\'a} for useful discussions. Jacob Ratkiewicz provided assistance in gathering and analyzing the Wikipedia dataset. This work was supported in part by a Lagrange Senior Fellowship of the CRT Foundation to FM, by the Institute for Scientific Interchange Foundation, the EPFL Laboratory of Statistical Biophysics, and the Indiana University School of Informatics; M.~A.~S. acknowledges support by DGES grant No.
FIS2007-66485-C02-01.
\end{acknowledgments}

\appendix
\section{Web Datasets}

We use three different datasets. The Industry Sector database (IS) is a collection of almost 10,000 corporate Web pages organized into 12 categories or sectors. The second dataset is a sample of the Open Directory (\url{dmoz.org}, ODP), a collection of Web pages classified into a large hierarchical taxonomy by volunteer editors. While the full ODP includes millions of pages, our collection comprises of approximately 150,000 pages, sampled uniformly from all top-level categories and crawled from the Web. The third corpus is a random sample of 100,000 topic pages from the English Wikipedia (\url{en.wikipedia.org}, Wiki), a popular collaborative encyclopedia which also is comprised of millions of online entries.

\begin{figure}[h]
\centerline{\includegraphics[width=0.45\textwidth]{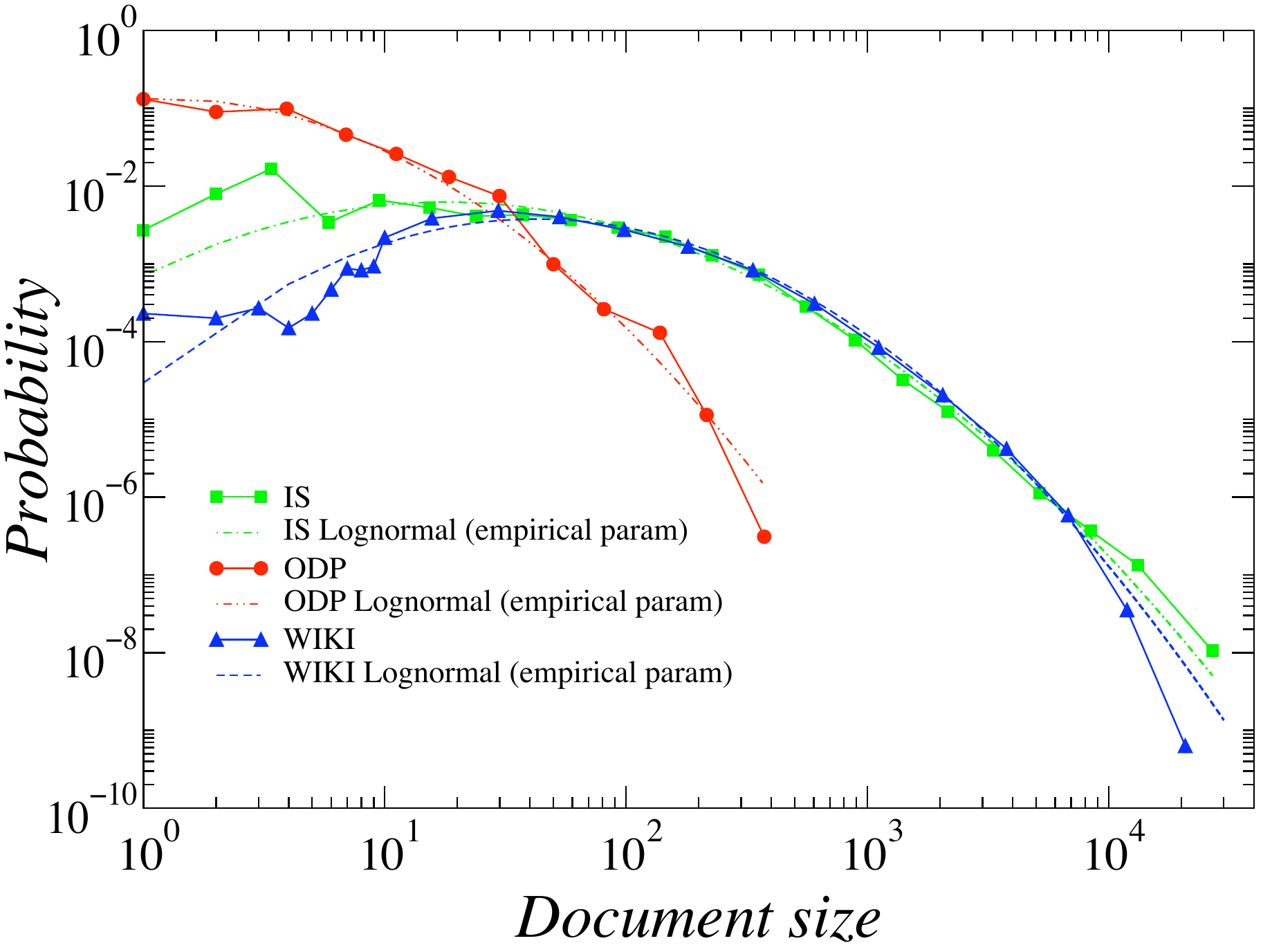}}
 \caption{Distributions of document length for all three collections. Each is very well approximated by a log-normal, with different first and second moment parameters (see Table~\ref{tableDB}).}
\label{lognormal}
\end{figure}

These English text collections are derived from public data and are publicly available (IS dataset is available at \url{www.cs.umass.edu/~mccallum/code-data.html}, ODP and Wikipedia corpora available
upon request); have been used in several previous studies, allowing a cross check of our results; and are large enough for our purposes without being computationally unmanageable. The datasets are however very diverse in a number of ways. The IS corpus is relatively small and topically focused, while ODP and Wikipedia are larger and have broader coverage, as reflected in their vocabulary sizes. IS documents represent corporate content, while many Web pages in the ODP collection are individually authored. Wikipedia topics are collaboratively edited and thus represent the consensus of a community. In spite of such differences, the distributions of document length for all three collections are very well approximated by log-normals shown in Fig.~\ref{lognormal}, with different first and second moment parameters. Table~\ref{tableDB} summarizes the main statistical features of the three collections. Before our analysis, all documents in each collection have been parsed to extract the text (removing HTML markup) and syntactic variations of words have been conflated using standard stemming techniques~\cite{Porter80}.

\begin{table}
\caption{Statistics for the different document collections. $V$ stands for vocabulary size, $D$ for the number of documents containing at least one word (in parenthesis the number of empty documents in the collection), $\langle w \rangle$ for the average size of documents in number of unique words, and $\langle n \rangle$ and $\sigma^2(n)$ for the average and variance of document size in number of words. For each collection, the distribution of document size is very well fitted by a lognormal with parameters $\mu$ and $\sigma^2$.}
\begin{center}
\scriptsize
\begin{tabular}{l|cccccccc}
\hline
Dataset & $V$ & $D$ & $\langle w \rangle$ & $\langle n \rangle$ & $\sigma^2(n)$ & $\mu$ & $\sigma^2$\\
\hline
Wiki & 588639 & 100000 (0) & 160.44 & 373.86 & 457083 & 5.20 & 1.45 \\
IS & 47979 & 9556 (15) & 124.26 & 313.46 & 566409 & 4.79 & 1.91\\
ODP & 105692 & 107360 (32558) & 8.88 & 10.34 & 345 & 1.62 & 1.44\\
\hline
\end{tabular}
\end{center}
\label{tableDB}
\end{table}

\section{Analytical derivation of Heap's law within our model}

The probability $P(w,n)$ to find $w$ different words in a document of size $n$ satisfies the following discrete master equation:
\begin{equation}
\label{master}
P(w+1,n+1)=P(w+1,n)F(w+1)+P(w,n)(1-F(w))
\end{equation}
where $F(w)=\sum_{r=1}^{w} P(r)$, and $P(r)$ is the Zipf probability associated with rank $r$. The tail of the Zipfian rank distribution is critical because the words not yet observed occupy the ranks at the bottom of the frequency distribution ($r>w$), and their cumulative probability is therefore $1-F(w)$. Multiplying both sides of Eq.~\ref{master} by $w+1$ and summing over $w$ leads to a relation between the expected values $E[w]$ of the number of different words for document sizes $n+1$ and $n$:
\begin{equation}
E[w(n+1)]=E[w(n)]+E[ 1-F(w(n)) ]
\label{relation}
\end{equation}
where the second term in the {\it r.h.s.} states that the probability to observe a new word (when $w$ different words are already present in the document) is the cumulative probability of words with frequency ranking larger than $w$. Neglecting fluctuations and taking the continuous limit, Eq.~\ref{relation} leads to
\begin{equation}
\frac{dw(n)}{dn}=\int_{w(n)}^{V} P(r)dr.
\label{eq:heap}
\end{equation}
Eq.~\ref{eq:heap} can be integrated numerically using the actual $P(r)$ from the data. Alternatively, Eq.~\ref{eq:heap} can be solved analytically for special cases (see main text).

\section{Algorithm}

The dynamic ranking model is implemented by the following algorithm:
%\begin{figure}
\vspace{-0.5cm}
\begin{alltt} \baselineskip18pt
Vocabulary: \(t \in \{1, \ldots, V\}\)
Initial ranking: \(\forall t: r\sb{0}(t)=t\)
Repeat until \(D\) documents are generated:
   Initialize term counts to \(\forall t: c(t)=0\) (*)
   Draw \(L\) from lognormal(\(\mu, \sigma\sp2\))
   Repeat until \(L\) terms are generated:
      Sort terms to obtain new rank \(r(t)\)
         according to \(c(t)\) (break ties by \(r\sb0\))
      Select term \(t\) with probability \(P(t)\propto r(t)\sp{-1}\)
      Add \(t\) to current document
      \(c(t) \leftarrow c(t) + 1\)
   End of document
End of collection
\end{alltt}
%\caption{Pseudocode of the dynamic ranking model. The document initialization step (line marked with an asterisk) is altered in the memory version of the model (see text).}
%\label{algo:drof}
%\end{figure}

The document initialization step (line marked with an asterisk in above pseudocode) is altered in the more general, memory version of the model (see main text). In particular we set to zero the counts $c(t)$ not of all terms, but only of terms $t$ such that $r(t) \geq r^{*}$. The rank $r^{*}$ is drawn from an exponential distribution $P(r^*) = z(1-z)^{r^* - 1}$ where $z$ is a probability parameter that measures the lexical diversity of the collection and $r^*$ has expected value $1/z$. In simpler terms, the counts of the $r^*$ top-ranked words are preserved while all the others are reset to zero.

Algorithmically, terms are sorted by counts so that the top-ranked term $t$ ($r(t)=1$) has the highest $c(t)$. We iterate over the ranks $r$, flipping a biased coin for each term. As long as the coin returns false (probability $1-z$), we preserve $c(t(r))$. As soon as the coin returns true (probability $z$), say for the term $t(r^*)$, we reset all the counts for this and the following terms: $\forall r \geq r^* \;\; c(t(r))=0$.

The special case $z=1$ reverts to the original, memory-less model; all counts are reset to zero and each document restarts from the global Zipfian ranking $r_0$. The special case $z=0$ is equivalent to the Zipf null model as the term counts are never reset and thus rapidly converge to the global Zipfian frequency distribution.

%\bibliography{zipf}
%\bibliographystyle{apsrev}

\end{document}